\documentclass{bmvc2k}

\usepackage[dvipsnames]{xcolor,colortbl}         
\usepackage[utf8]{inputenc} 
\usepackage[T1]{fontenc}    
\usepackage{url}            
\usepackage{booktabs}       
\usepackage{amsfonts}       
\usepackage{nicefrac}       
\usepackage{comment}

\usepackage{multirow}
\usepackage{capt-of}
\usepackage{microtype}

\usepackage{enumitem}

\usepackage{threeparttable} 

\usepackage{booktabs}

\usepackage{wrapfig}
\definecolor{OursColor}{HTML}{F2F3F4}

\newcommand{\ul}[1]{{\underline{#1}}}

\newcommand{\ourso}{\texttt{GeoMAD} (ours)\xspace}
\newcommand{\ours}{\texttt{GeoMAD}\@\xspace}

\newcommand{\ie}{\textit{i.e.}\ }
\newcommand{\etc}{\textit{etc.}\ }

\newcommand{\submission}{}

\ifx \submission \undefined
\definecolor{DarkGreen}{rgb}{0.00, 0.40, 0.00}
\definecolor{Red}{rgb}{1.0, 0.0, 0.0}

\newcommand{\SC}[1]{\textcolor{MidnightBlue}{#1}}
\else

\newcommand{\SC}[1]{{#1}}
\fi

\def\gG{{\mathcal{G}}}

\usepackage{pifont}
\newcommand{\xmark}{\ding{55}}
\definecolor{checkmark}{HTML}{305AFF}
\definecolor{xmark}{HTML}{E62020}
\newcommand{\ccheck}{{\textcolor{checkmark}{\large\checkmark}}}
\newcommand{\ccross}{{\textcolor{xmark}{\large\xmark}}}

\title{GeoMAD: Geometry-Aware Multi-View Anomaly Detection via Deformable Fusion and Distributional Alignment}

\addauthor{Shang-Fu Chen}{chenshangfu@cmlab.csie.ntu.edu.tw}{1}
\addauthor{Jhih-Ciang Wu}{jcwu@csie.ntnu.edu.tw}{2}
\addauthor{Kuan-Chuan Peng}{kpeng@merl.com}{3}
\addauthor{Wen-Huang Cheng}{wenhuang@csie.ntu.edu.tw}{1,5}
\addauthor{Kai-Lung Hua}{kai.hua@microsoft.com}{4,6}

\addinstitution{
 National Taiwan University\\
 Taipei, Taiwan
}
\addinstitution{
 National Taiwan Normal University\\
 Taipei, Taiwan
}
\addinstitution{
 Mitsubishi Electric Research Laboratories (MERL)\\
 Cambridge, MA, USA
}
\addinstitution{
 Microsoft Taiwan Corporation\\
 Taipei, Taiwan
}
\addinstitution{
 VinUniversity\\
 Hanoi, Vietnam
}
\addinstitution{
 National Taiwan University of Science and Technology\\
 Taipei, Taiwan
}

\runninghead{Chen et al.}{Geometry-Aware Multi-view Anomaly Detection}

\begin{document}

\maketitle

\begin{abstract}
Multi-view anomaly detection (MvAD) detects defects by exploiting complementary observations from multiple camera viewpoints. The central challenge is to fuse views with sufficient geometric awareness while remaining scalable to multi-class industrial settings. Existing methods typically fall into two extremes: voxel-based fusion provides explicit geometric alignment but requires costly 3D construction and class-specific assumptions, whereas lightweight patch-based fusion is efficient but relies on discrete candidate matching and lacks continuous cross-view correspondence. In this paper, we propose \ours, a unified multi-view, multi-class AD framework that addresses both geometric correspondence deficiency and distributional inconsistency. \SC{Our \textit{Cross-view Deformable Fusion Module} (CDFM) learns content-adaptive, view-pair-specific sampling offsets directly on 2D feature maps and arranges them across a multi-scale window pyramid with image-global reference sampling}, enabling hierarchical cross-view correspondence without camera calibration, voxel construction, or class-specific 3D supervision. We further introduce \textit{Distributional View Alignment} (DVA), a self-supervised cross-view regularization loss that aligns each view's bottleneck distribution against a per-instance view-centric target, enforcing global consistency without pixel-level correspondence. Together, CDFM and DVA bridge local geometric correspondence and global distributional consistency, providing geometry-aware and distribution-consistent fusion while preserving the efficiency of 2D feature-space learning. Extensive experiments on Real-IAD and MANTA-Tiny show that \ours achieves strong detection and localization performance in unified MvAD.
\end{abstract}

\section{Introduction}
\label{sec:introduction}

Anomaly Detection (AD) is a fundamental task in industrial inspection~\cite{pang2021deep,miyai2024generalized,zhang2025systematic,lin2025survey}, medical imaging~\cite{cai2025medianomaly,zhang2024mediclip,cai2024rethinking}, and video surveillance~\cite{liu2025networking,zhu2024advancing,li2025anomize,majhi2025just}, where the goal is to identify deviations from normal patterns before they lead to costly failures or safety risks. 
Since abnormal samples are rare and diverse, most AD methods~\cite{bergmann2019mvtec,zou2022spot,ruff2020deep} are trained using only normal data and detect anomalies by measuring deviations from learned normality. Existing one-class methods~\cite{ruff2018deep,tien2023revisiting,chen2025filter,liu2025unlocking,nafez2025patchguard,zhang2024realnet,liu2023simplenet} often train a separate model for each object class. Although effective, this design becomes expensive to deploy across multiple product lines. Recent unified AD methods~\cite{guo2025dinomaly,wei2025uninet,zhang2025exploring} address this issue by training a single model across multiple classes, but they still largely operate under a single-view assumption.

\begin{figure}[t]
    \centering
    \includegraphics[width=1\linewidth]{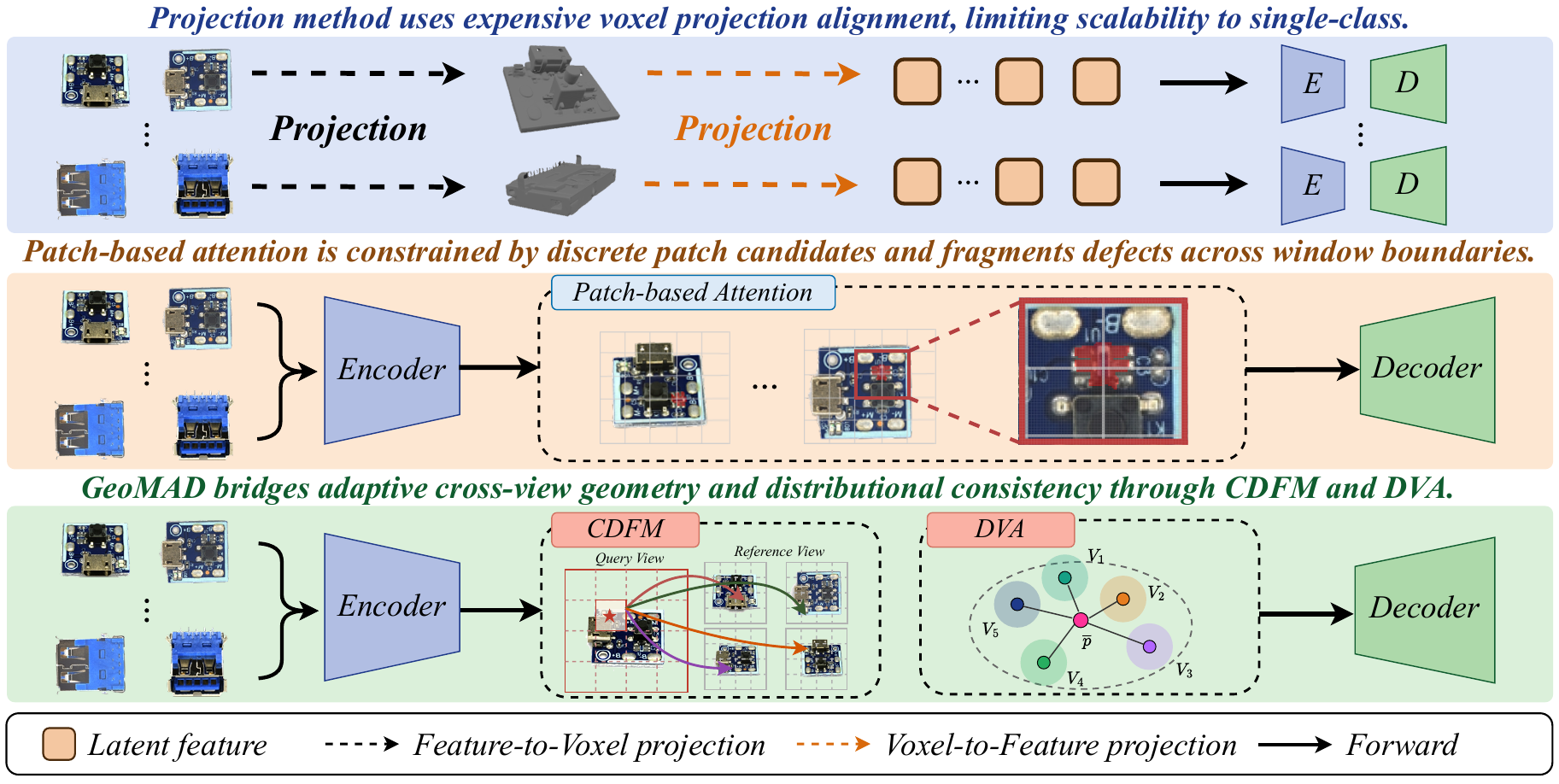}
    \vspace{-2em}
    \caption{
    \ours combines Cross-view Deformable Fusion Module (CDFM) and Distributional View Alignment (DVA) to achieve geometry-aware and distribution-consistent fusion directly in 2D feature space, addressing key limitations of existing methods~\cite{he2024learning, mao2025unveiling}.
    }
    \vspace{-1em}
    \label{fig:teaser}
\end{figure}


Single-view imagery is inherently limited for anomaly detection. Defects may be occluded, weakly illuminated, or geometrically ambiguous from one camera angle, while being clearly visible from another. Multi-view anomaly detection (MvAD)\footnote{We use MvAD to denote the Multi-view Anomaly Detection task, while MVAD refers to a previous work~\cite{he2024learning}.} addresses this limitation by leveraging multiple synchronized observations of the same object. Given views $\{X_1,\ldots,X_V\}$, MvAD aims to exploit their complementary information for more reliable anomaly detection and localization than any single view. Recent multi-view datasets such as Real-IAD~\cite{wang2024real} and MANTA~\cite{fan2024manta} have made this setting increasingly practical. However, effective MvAD requires more than simply processing multiple images. The key challenge is how to fuse views in a way that is both geometrically meaningful and computationally scalable.
Existing cross-view fusion strategies expose a clear trade-off between geometric fidelity and computational scalability, as illustrated in Figure~\ref{fig:teaser}. Voxel-projection methods~\cite{mao2025unveiling} establish explicit cross-view correspondence by lifting features into a shared 3D representation. This provides strong geometric alignment, but requires expensive voxel construction and is often tied to class-specific 3D assumptions. Patch-attention methods~\cite{he2024learning}, by contrast, are lightweight and more general, but their correspondence is still mediated by discrete patch-level candidate selection. As a result, they lack an adaptive mechanism for cross-view geometry and may fragment anomalies that cross patch or window boundaries. These limitations are particularly problematic for MvAD, where the same surface defect can appear at different image locations, scales, and shapes across views.

We identify two issues that an effective multi-view AD model should address. The first is \textit{geometric correspondence deficiency}: lightweight patch-based fusion can retrieve cross-view candidates through patch-level similarity and then refine them with pixel-wise attention, but the correspondence remains constrained by discrete patch candidates rather than continuous geometric displacement. Such coarse candidate selection provides limited adaptability to viewpoint-dependent changes in location, scale, and shape, and may still fragment a continuous defect when its corresponding evidence is distributed across multiple patch regions. The second is \textit{distributional inconsistency}: even after local cross-view information is exchanged, the fused representations from different views are not explicitly encouraged to share compatible normality statistics. Pixel-level consistency is not suitable for resolving this issue, because the same physical surface region may appear at different spatial locations across cameras. Instead, multi-view AD requires both adaptive local geometric correspondence and global distributional consistency.

\SC{
To address these challenges, we propose \ours, a unified multi-view, multi-class AD framework built upon reverse distillation. \ours introduces \textit{Cross-view Deformable Fusion Module} (CDFM), a bottleneck design that learns content-adaptive, view-pair-specific sampling offsets with image-global reference reach across a hierarchical window pyramid. By sampling reference-view features from continuous, content-adaptive locations at multiple scales, CDFM provides implicit geometric reasoning without camera calibration, voxel construction, or class-specific 3D supervision.
We further introduce \textit{Distributional View Alignment} (DVA), a self-supervised cross-view regularization loss that aligns each view's bottleneck distribution against a per-instance view-centric target. The synchronized views of the same normal object provide a natural supervisory signal, allowing DVA to encourage compatible feature statistics without anomaly labels, pixel-level matching, or explicit 3D projection. CDFM learns local deformable correspondence, while DVA enforces global view-level consistency. Together, they enable geometry-aware and distribution-consistent multi-view AD without the memory and supervision requirements of explicit 3D alignment.
We evaluate \ours on Real-IAD~\cite{wang2024real} and MANTA-Tiny~\cite{fan2024manta}. Extensive experiments and ablation studies show that \ours achieves strong anomaly detection and localization performance in unified multi-view AD, while maintaining practical computational efficiency.
}

Our main contributions are summarized as follows:
\begin{itemize}[leftmargin=*, itemsep=0em]
\item We propose \ours, a unified multi-view, multi-class AD framework that jointly addresses adaptive cross-view correspondence and view-level distributional consistency without explicit 3D reconstruction.
\item \SC{We propose \textit{Cross-view Deformable Fusion Module} (CDFM), a calibration-free 2D bottleneck design that learns content-adaptive, view-pair-specific sampling offsets with image-global reach for geometry-aware multi-view fusion.}
\item We introduce \textit{Distributional View Alignment} (DVA), which regularizes cross-view consistency in distribution space via a per-instance view-centric target, without requiring pixel-level correspondence or explicit 3D projection.
\item Extensive experiments on Real-IAD and MANTA-Tiny demonstrate that \ours improves unified multi-view anomaly detection and localization over existing methods.
\end{itemize}

\section{Related Work}
\label{sec:related_work}

\begin{table}[!t]
\setlength{\tabcolsep}{4pt}
\centering
\resizebox{.76\columnwidth}{!}{
\begin{tabular}{lc@{\hspace{3mm}}c@{\hspace{3mm}}c@{\hspace{3mm}}c>{\columncolor{OursColor}}c}
\toprule
\multirow{2.5}{*}{\textbf{\shortstack{Method Properties}}}
& \multicolumn{5}{c}{\textbf{AD Categories}} \\
\cmidrule(l){2-6}
& $\gG_0$ & $\gG_1$ & IDIF & MVAD & \ourso \\
\midrule
Single model for all classes              & \ccross & \ccheck & \ccross & \ccheck & \ccheck \\
Designed for multi-view AD                & \ccross & \ccross & \ccheck & \ccheck & \ccheck \\
Learned 2D geometric correspondence       & \ccross & \ccross & \ccross & \ccross & \ccheck \\
Distributional view alignment             & \ccross & \ccross & \ccross & \ccross & \ccheck \\
\bottomrule
\end{tabular}
}
\caption{
Characteristic comparison between \ourso and prior AD works.
$\gG_0$ denotes class-specific single-view AD methods~\cite{liu2023simplenet,hu2024anomalydiffusion,roth2022towards,defard2021padim,deng2022anomaly,zavrtanik2021draem,chen2025filter};
$\gG_1$ denotes unified single-view AD methods~\cite{you2022unified,zhang2025exploring,he2024mambaad,guo2025dinomaly,he2024diffusion,wei2025uninet}.
IDIF~\cite{mao2025unveiling} and MVAD~\cite{he2024learning} denote representative multi-view AD methods.
}
\label{tab:related_works}
\end{table}

\noindent\textbf{Unsupervised anomaly detection:}
Unsupervised AD aims to identify anomalies and localize defects using only normal samples during training, assuming anomalies are rare and can be distinguished from learned normal patterns.
Existing AD methods can be broadly grouped into class-specific and unified settings, as summarized in Table~\ref{tab:related_works}.
Class-specific approaches $\gG_0$ train separate models for individual object categories, achieving strong performance but incurring high storage and deployment costs. Unified approaches $\gG_1$ use a single model across categories, improving scalability while introducing potential inter-class interference.
Category $\gG_0$ includes works~\cite{liu2023simplenet,hu2024anomalydiffusion,roth2022towards,defard2021padim,deng2022anomaly,zavrtanik2021draem,chen2025filter} that use separate models for each object class. These methods can be further organized into three styles, \ie, synthesis-based, embedding-based, and reconstruction-based approaches.
Synthesis-based methods~\cite{liu2023simplenet,hu2024anomalydiffusion} create artificial anomalies during training but suffer from synthetic-to-real domain gaps and fail to capture cross-view spatial relationships.
Embedding-based methods~\cite{roth2022towards,defard2021padim} encode images into feature representations using pre-trained networks; however, they encounter feature misalignment issues in multi-view settings where features from different viewpoints lack spatial correspondence.
Reconstruction-based methods~\cite{deng2022anomaly,chen2025filter,zavrtanik2021draem} train encoder-decoder networks to reconstruct normal patterns, but may inadvertently reconstruct abnormal regions, leading to false negatives that are exacerbated in multi-view contexts without proper cross-view constraints.
Category $\gG_1$ encompasses recent works~\cite{you2022unified,zhang2025exploring,he2024mambaad,guo2025dinomaly,he2024diffusion,wei2025uninet} that require only a single model for all object classes. ViTAD~\cite{zhang2025exploring} utilizes Vision Transformers for feature regression, while MambaAD~\cite{he2024mambaad} employs state space models for both global and local modeling. While these unsupervised methods have shown efficacy in unified scenarios~\cite{lv2025oneforall}, they typically struggle with multi-view designs, often resorting to single-view voting ensembles that process each view independently and yield suboptimal sample-level scores. Compared to these previous methods, we propose a unified architecture that carries out multi-view AD while explicitly exploring correlations across different viewpoints.


\noindent\textbf{Multi-view anomaly detection:}
Building on single-view AD, recent works address the harder multi-view setting. MVAD~\cite{he2024learning} uses patch-based cross-attention for adaptive multi-view selection, but suffers from \textit{geometric correspondence deficiency}: patch partitioning fragments boundary anomalies, causing incomplete detection. It also lacks cross-view constraints, yielding \textit{distributional inconsistency} where representations from different views fail to share compatible normality statistics. IDIF~\cite{mao2025unveiling} aligns views via voxel projection, but requires costly voxel transformations and is limited to single-class settings, as diverse geometries hinder unified 3D modeling. As shown in Table~\ref{tab:related_works}, \ours unifies learned 2D cross-view correspondence via CDFM and distributional view alignment via DVA, avoiding explicit 3D reconstruction while supporting multi-class AD.

\section{Our proposed Method -- \ours}
\label{sec:method}

\begin{figure*}[t]
    \centering
    \includegraphics [width=.985\textwidth]{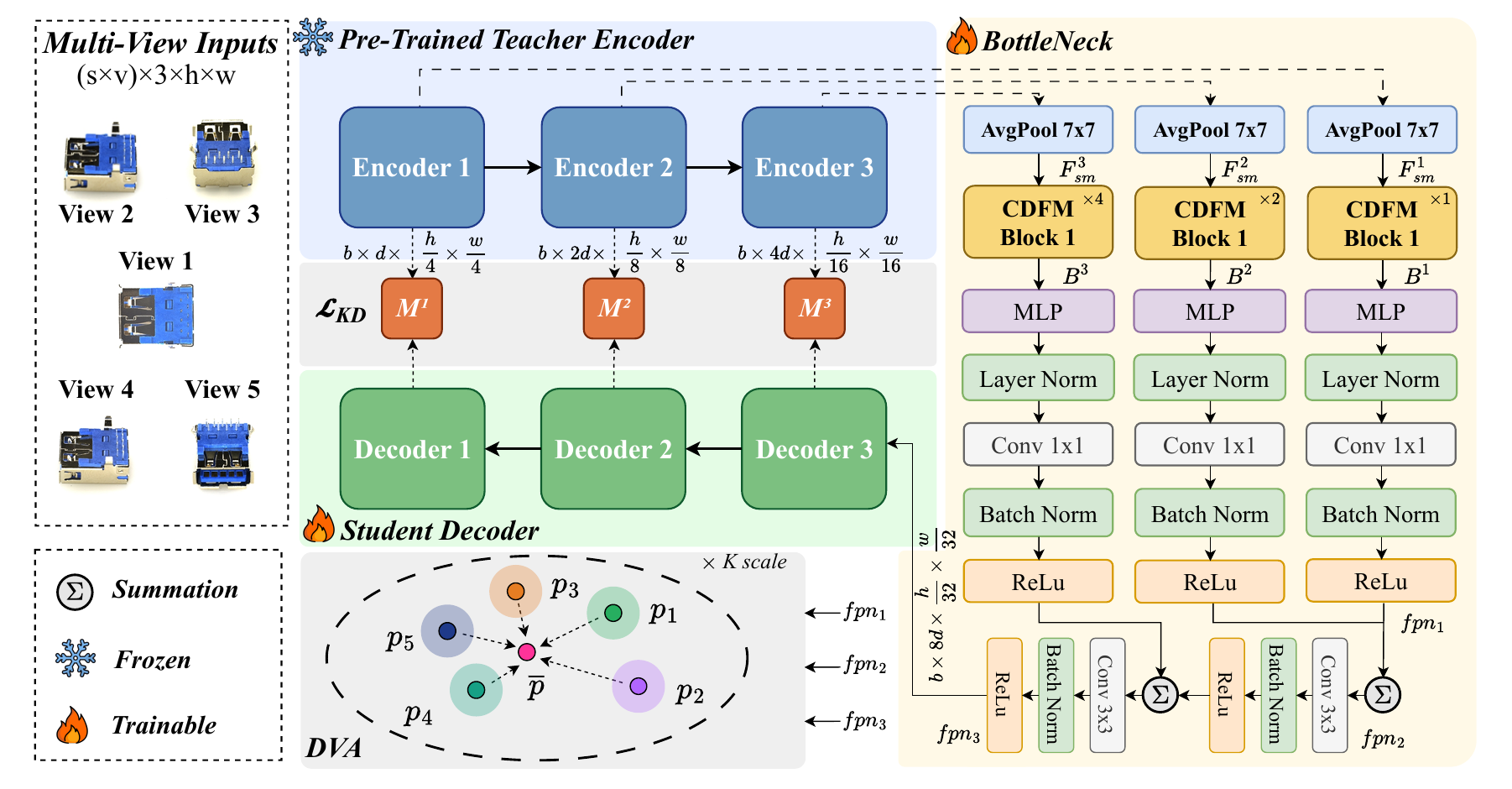}
    \vspace{-1em}
    \caption{
    \SC{
    Overview of our proposed \ours. Given multi-view input images, \ours adopts a teacher-student architecture with the Cross-view Deformable Fusion Module (CDFM) for deformable cross-view feature fusion. Distributional View Alignment (DVA) mitigates distributional inconsistency by aligning bottleneck feature distributions across views. Multi-view anomaly scores are then aggregated for final detection.}
    }
    \label{fig:framework}
\end{figure*}

\subsection{Overview}
\label{subsec:overview}

Let $\mathcal{D}_{\mathrm{train}}=\{\mathcal{I}_1,\ldots,\mathcal{I}_n\}$ denote a set of $n$ anomaly-free multi-view training instances, where each instance $\mathcal{I}_i=\{I_i^1,\ldots,I_i^V\}$ contains $V$ views of the same object. Let $\mathcal{D}_{\mathrm{test}}=\{\mathcal{I}_{n+1},\ldots,\mathcal{I}_{n+m}\}$ denote a test set containing both normal and anomalous instances. The goal is to learn, using only anomaly-free training data, a model that can detect and localize anomalies in the test set.
Following~\cite{he2024learning, mao2025unveiling}, we formulate multi-view AD by organizing each sample as a set of synchronized views. Unlike single-view AD methods that process a batch of $B$ independent images, multi-view methods restructure the batch as $B=S\times V$, where $S$ is the number of object instances and $V$ is the number of views per instance. This organization enables cross-view interaction within each object instance while preserving efficient batched computation.
As illustrated in Figure~\ref{fig:framework}, \ours is designed for multi-view, multi-class AD. Following the reverse distillation paradigm of RD4AD~\cite{deng2022anomaly}, \ours consists of three core components: a pretrained teacher network, a trainable student network, and a cross-view bottleneck module based on our proposed \textit{Cross-view Deformable Fusion Module} (CDFM).

Given a multi-view input $\mathbf{X}\in\mathbb{R}^{B\times 3\times h\times w}$, \ours processes each view through the teacher and student networks. The frozen teacher extracts multi-scale features $\mathbf{T}^k\in\mathbb{R}^{B\times d_k\times h_k\times w_k}$, where $k\in\{1,\ldots,K\}$ indexes the feature scale. The CDFM bottleneck reshapes these features into multi-view form, $\mathbb{R}^{S\times V\times d_k\times h_k\times w_k}$, and performs cross-view fusion at each scale to produce fused bottleneck representations $\mathbf{B}^k$. These multi-scale representations are merged through a top-down feature-pyramid fusion into a single coarsest-scale feature, which conditions the trainable student decoder that reconstructs multi-scale features $\mathbf{S}^k\in\mathbb{R}^{B\times d_k\times h_k\times w_k}$ matching the teacher's scales.
The training objective combines the multi-scale reverse-distillation loss $\mathcal{L}_{\mathrm{KD}}$ with the proposed Distributional View Alignment loss $\mathcal{L}_{\mathrm{DVA}}$ (Sec.~\ref{subsec:DVA}), which regularizes view-level feature statistics within each multi-view instance. $\mathcal{L}_{KD}$ measures feature alignment between the teacher and student networks across $K$ scales:
\begin{equation}
\mathcal{L}_{KD} = \sum_{k=1}^{K} \left\{ \frac{1}{H_k W_k} \sum_{h=1}^{H_k} \sum_{w=1}^{W_k} M^k(h, w) \right\},
\end{equation}
where $M^k(h, w)$ is the cosine distance between the teacher and student features at spatial location $(h, w)$ and scale $k$. The multi-scale formulation captures anomalies at different semantic and spatial resolutions, while $\mathcal{L}_{\mathrm{DVA}}$ encourages distributional consistency across views of the same object instance.

At inference, \ours generates dense pixel-level anomaly scores $\mathbf{A}_{px}\in\mathbb{R}^{mV\times H\times W}$ by aggregating the multi-scale cosine distances $M^k(y,x)$ between teacher and student features. The image-level anomaly score $\mathbf{A}_{im}\in\mathbb{R}^{mV}$ is obtained by spatial max-pooling over each view. The final sample-level anomaly score $\mathbf{A}_{sa}\in\mathbb{R}^{m}$ is computed by first applying spatial max-pooling across each view, then averaging across all $V$ viewpoints within each sample, effectively capturing the most abnormal perspective while leveraging multi-view consensus.

\subsection{Cross-view Deformable Fusion Module}
\label{subsec:CDFM}

\begin{figure*}[t]
    \centering
    \includegraphics [width=.985\textwidth]{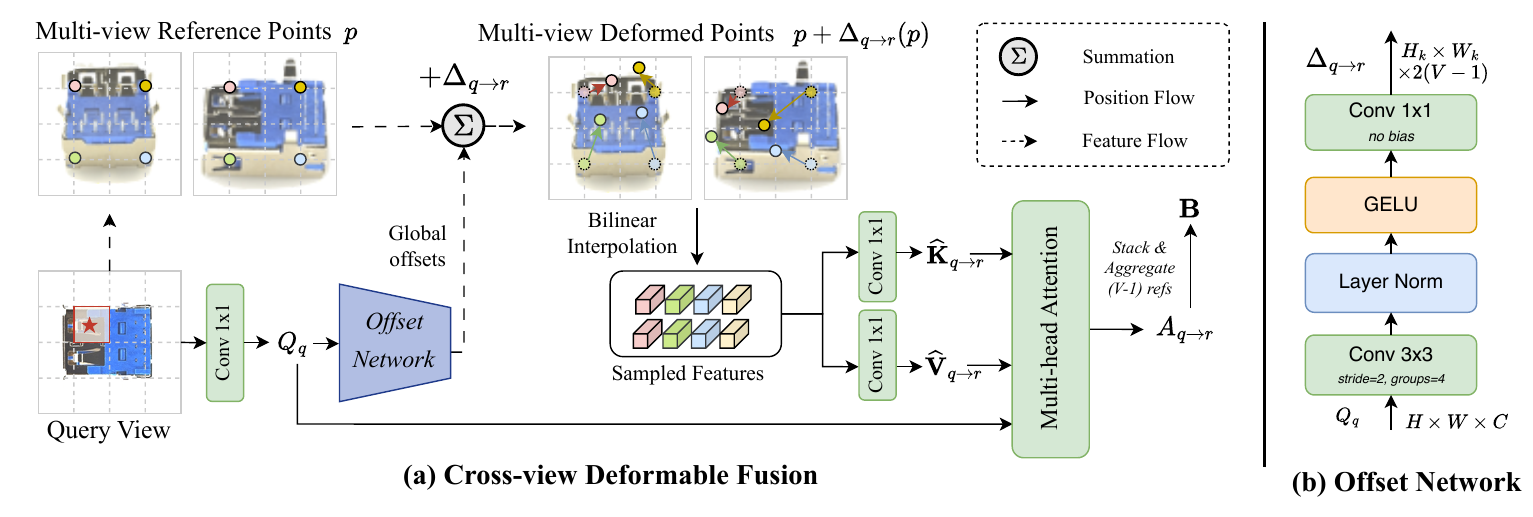}
    \vspace{-1em}
    \caption{
    \SC{
    The mechanism of CDFM. (a) CDFM predicts offsets $\boldsymbol{\Delta}_{q\to r}$ from the windowed query features and bilinearly samples reference keys and values at the deformed positions $\mathbf{p} + \boldsymbol{\Delta}_{q\to r}$. Each query view produces $V{-}1$ offset fields, so windowed queries can deformably attend to features anywhere in each reference view, coupling cheap query windowing with global key/value reach. (b) Details of the offset network.
    }
    }
    \label{fig:cdfm}
\end{figure*}

Multi-view AD requires cross-view interaction that is both geometrically meaningful and computationally efficient. Existing fusion strategies usually satisfy only one side of this requirement. Voxel-projection methods establish explicit cross-view correspondence by lifting features into a shared 3D representation, but this introduces expensive voxel construction and often depends on object-specific 3D priors. Patch-attention methods, in contrast, avoid 3D construction and remain lightweight, but their fixed partitioning provides no explicit mechanism for cross-view geometry and may fragment defects that cross patch boundaries.
We address this limitation with \textit{Cross-view Deformable Fusion Module} (CDFM), a compact bottleneck that aligns multi-view features directly in 2D feature space, as shown in Figure~\ref{fig:cdfm}. Given features from multiple views, CDFM learns data-dependent sampling offsets that specify where each query location should gather information from the remaining views. These offsets serve as an implicit geometric prior, enabling cross-view correspondence without camera calibration, voxel construction, or class-specific 3D supervision. Compared with fixed patch-wise interaction, CDFM introduces only a small overhead from offset prediction and bilinear sampling while providing geometry-aware deformable fusion.

For clarity, we describe CDFM at a single feature scale and omit the scale index $k$ when no ambiguity arises. The same operation is applied independently at all selected feature scales. Let $\mathbf{T}\in\mathbb{R}^{B\times C\times H\times W}$ denote the teacher feature at the current scale, where $B=S\times V$. We reshape it into multi-view form $\mathbf{F}\in\mathbb{R}^{S\times V\times C\times H\times W}$. We use $q\in\{1,\ldots,V\}$ to index the query view and $r\neq q$ to index a reference view.


\noindent\textbf{Within-view spatial smoothing.}
Before cross-view fusion, we apply a parameter-free average pooling operation to reduce local discontinuities in the bottleneck features:
\begin{equation}
\label{eq:smoothing}
\mathbf{F}_{\mathrm{sm}}
=
\mathrm{AvgPool}_{7\times7}(\mathbf{F}).
\end{equation}
Here, $\mathrm{AvgPool}_{7\times7}$ denotes channel-wise average pooling with kernel size $7\times7$, stride $1$, and padding $3$, which preserves the spatial resolution. The operation is applied independently to each view and channel, introduces no learnable parameters, and smooths local boundary artifacts while preserving surface structure. 
\SC{We use the same kernel size at all feature scales and analyze this choice in Table~\ref{tab:ablation_kernel_size}.}

\noindent\textbf{Deformable cross-view attention.}
Given the smoothed multi-view feature $\mathbf{F}_{\mathrm{sm}}$, CDFM treats each view as the query view in turn and uses the remaining views as references. The goal is to let each query location adaptively gather information from geometrically corresponding regions in other views, instead of attending to fixed patch windows. For a query view $q$ and a reference view $r\neq q$, we compute the query, key, and value projections:
\begin{equation}
\label{eq:qkv}
\mathbf{Q}_{q}
=
\mathrm{Conv}_{q}(\mathbf{F}_{\mathrm{sm},q}),
\quad
\mathbf{K}_{r}
=
\mathrm{Conv}_{k}(\mathbf{F}_{\mathrm{sm},r}),
\quad
\mathbf{V}_{r}
=
\mathrm{Conv}_{v}(\mathbf{F}_{\mathrm{sm},r}),
\end{equation}
where $\mathrm{Conv}_{q}$, $\mathrm{Conv}_{k}$, and $\mathrm{Conv}_{v}$ are $1\times1$ convolutions for query, key, and value projection, respectively.
To establish cross-view correspondence, CDFM predicts sampling offsets from the query feature rather than using fixed reference locations. A lightweight offset predictor generates one two-channel offset field for each reference view:
\begin{equation}
\label{eq:offset}
\boldsymbol{\Delta}_{q}
=
\mathrm{ConvOff}(\mathbf{Q}_{q})
\in
\mathbb{R}^{S\times 2(V-1)\times H\times W},
\quad
\boldsymbol{\Delta}_{q\to r}
=
\boldsymbol{\Delta}_{q}[r].
\end{equation}
Here, $\mathrm{ConvOff}$ is composed of a depthwise spatial convolution followed by a pointwise $1\times1$ convolution. Its output is divided into $V-1$ two-channel offset fields, where $\boldsymbol{\Delta}_{q\to r}(p)\in\mathbb{R}^{2}$ denotes the learned displacement from query view $q$ to reference view $r$ at location $p$ on the offset grid. Since different camera pairs induce different geometric displacements, predicting separate offsets for each reference view allows CDFM to model view-pair-specific correspondence while sharing the same offset predictor.

Using these offsets, reference keys and values are sampled at deformed positions by bilinear interpolation:
\begin{equation}
\label{eq:sample}
\hat{\mathbf{K}}_{q\to r}(p)
=
\mathbf{K}_{r}
\left(
p+\boldsymbol{\Delta}_{q\to r}(p)
\right),
\quad
\hat{\mathbf{V}}_{q\to r}(p)
=
\mathbf{V}_{r}
\left(
p+\boldsymbol{\Delta}_{q\to r}(p)
\right).
\end{equation}
This deformable sampling step lets each query location retrieve reference-view features from continuous, content-adaptive positions, which alleviates the boundary fragmentation caused by rigid patch partitioning.
The sampled keys and values are then used to compute the cross-view attention response:
\begin{equation}
\label{eq:pair_attention}
\mathbf{A}_{q\to r}
=
\mathrm{softmax}
\left(
\frac{
\mathbf{Q}_{q}
\left(\hat{\mathbf{K}}_{q\to r}\right)^{\top}
}
{\sqrt{d_h}}
\right)
\hat{\mathbf{V}}_{q\to r},
\end{equation}
\SC{
where $d_h$ is the per-head channel dimension. Finally, the fused bottleneck representation for each query view $q$ is obtained by aggregating the responses from all reference views, and stacking across query views yields the full fused bottleneck at this scale:
\begin{equation}
\label{eq:cdfm}
\mathbf{B}_{q} = \frac{1}{V-1} \sum_{r\neq q} \mathbf{A}_{q\to r},
\quad
\mathbf{B} = \mathrm{stack}_{q=1}^{V}\,\mathbf{B}_{q} \in \mathbb{R}^{S\times V\times C\times H\times W}.
\end{equation}
We use uniform aggregation as a view-symmetric prior, since each camera provides a complementary observation of the same object surface. The learned offsets determine where each reference view contributes information, while the averaging step avoids introducing an additional view-selection module that may overfit to easy views or suppress difficult but informative ones.
}

\noindent\textbf{Multi-scale geometric pyramid with global key-value reach.}
\SC{
CDFM organizes the query side into a multi-scale pyramid: at each feature scale, we partition the query feature map into $n_{\mathrm{win}}\times n_{\mathrm{win}}$ non-overlapping windows and predict offsets within each window from the local query features. We set $n_{\mathrm{win}}=\{8,4,2\}$ across the three feature scales, keeping the query window at a fixed $8\times8$ token size while letting its image-level receptive field grow from fine to coarse.
Although query computation is window-partitioned for efficiency, the key/value grid covers the entire reference feature map. Sampling positions are expressed in image-normalized coordinates rather than window-local coordinates, so a query in any window can deformably attend to reference features anywhere in the reference view. This decoupled query-window / global-key-value design retains the cost benefit of windowed queries, $\mathcal{O}(|\mathrm{window}|^2 \cdot n_{\mathrm{windows}})$ instead of $\mathcal{O}((HW)^2)$ per view pair. At the same time, it avoids the artificial boundaries between windows that pure window partitioning would create, which would otherwise block correspondence across views.}

\subsection{Distributional View Alignment}
\label{subsec:DVA}
CDFM learns local cross-view correspondence through deformable sampling, but local attention alone does not explicitly constrain the fused representations from different views to remain statistically consistent. We refer to this gap as \textit{distributional inconsistency}: locally fused features may still exhibit misaligned global statistics across views. This matters for multi-view AD because different views of the same normal object can have different spatial layouts, scales, and appearances, yet should preserve compatible normality statistics. Enforcing pixel-level consistency is therefore inappropriate, since the same surface region may appear at different image locations across cameras. Instead, we regularize cross-view consistency in distribution space.
We introduce \textit{Distributional View Alignment} (DVA), a view-level regularization loss applied to the CDFM bottleneck features. DVA aligns the feature distributions of different views within the same object instance, encouraging view-consistent bottleneck representations without interfering with the teacher-student distillation objective and without requiring the memory or supervision of explicit 3D alignment.

Concretely, we align every view's bottleneck distribution against a per-instance \emph{view-centric target} formed from the views themselves. This shared-target formulation avoids a conflicting-gradient pathology of more direct pairwise alignment, where one badly-aligned pair can dominate the gradient and pull two views apart along an axis that other pairs are simultaneously aligning. A single centroid target removes this contention.
Given bottleneck features $\mathbf{B}^k$ at scale $k$, let $\mathbf{b}^k_v$ denote the feature of view $v$. We first form the view-centric target $\bar{\mathbf{b}}^k$ by averaging across the $V$ views within the same object instance:
\begin{equation}
\label{eq:view_centroid}
\bar{\mathbf{b}}^k = \frac{1}{V} \sum_{v=1}^{V} \mathbf{b}^k_v.
\end{equation}
\SC{
We treat $\bar{\mathbf{b}}^k$ as a fixed target by stopping its gradient, denoted $\mathrm{sg}[\cdot]$ in Eq.~\ref{eq:dvaloss}. Without this, the averaging operation would propagate gradient from every view back into the centroid, creating a self-referential signal in which each view simultaneously shapes and is pulled toward its own target.}
We then convert each view feature and the centroid into probability distributions over the channel dimension via Softmax:
\begin{equation}
\label{eq:softmax_view}
p^k_v = \mathrm{softmax}_C(\mathbf{b}^k_v),
\quad
\bar{p}^k = \mathrm{softmax}_C(\bar{\mathbf{b}}^k).
\end{equation}
Distributional alignment is then enforced by the one-directional KL divergence of each view against the centroid, summed across views and scales:
\begin{equation}
\label{eq:dvaloss}
\mathcal{L}_{DVA} = \sum_{k=1}^{K} \sum_{v=1}^{V} D_{KL}\!\left( p^k_v \,\big\Vert\, \mathrm{sg}[\,\bar{p}^k\,] \right),
\quad
D_{KL}(p \,\Vert\, q) = \sum_{c} p(c) \log\frac{p(c)}{q(c)}.
\end{equation}
\SC{
where $K$ and $V$ denote the number of scales and views, respectively. The centric formulation is structurally symmetric across views (every view contributes equally to $\bar{p}^k$), so the asymmetric KL direction $D_{KL}(p^k_v \Vert \bar{p}^k)$ carries the design intent: it encourages each view to cover the modes of the consensus distribution rather than the consensus to cover any single view, which is the appropriate direction for normality modeling where capturing typical structure matters more than chasing outlier modes.}

\noindent\textbf{Assumption and scope.} \SC{DVA constrains the cross-fusion bottleneck only during training; at inference, anomaly maps are derived from per-view teacher--student feature divergence, which DVA does not modify. Single-view anomaly evidence in the frozen teacher features therefore remains intact, ensuring that view-dependent anomalies are still recoverable through the distillation path.
}

\noindent\textbf{Training objective.} Our final training objective combines reconstruction fidelity with distributional consistency:
\begin{equation}
\label{eq:total_loss}
\mathcal{L}_{total} = \mathcal{L}_{KD} + \lambda \mathcal{L}_{DVA},
\end{equation}
where $\mathcal{L}_{KD}$ is the knowledge-distillation loss and $\lambda$ is set to 1 to balance the two terms. $\mathcal{L}_{DVA}$ complements CDFM: CDFM establishes local cross-view correspondence through deformable sampling, while $\mathcal{L}_{DVA}$ enforces global distributional consistency on the resulting bottleneck. Together they yield a representation that is locally aligned and globally view-consistent without explicit 3D supervision.

\section{Experiments}
\label{sec:experiments}
\subsection{Experimental Settings}

\textbf{Datasets.} \textbf{Real-IAD}~\cite{wang2024real} is a large-scale multi-view industrial AD dataset containing 30 object classes across diverse materials, including metal, plastic, and mixed materials, \etc The dataset comprises 150K high-resolution images with 99,721 normal and 51,329 abnormal samples. The training and testing sets follow standard splitting protocols, with normal samples used for training, and both normal and abnormal samples for testing. The dataset features eight distinct defect types (pit, deformation, scratch, \etc) with defect areas ranging from 0.1\% to 6.75\%, making it significantly more challenging than existing benchmarks. \textbf{MANTA}~\cite{fan2024manta} is a visual-text AD dataset for tiny objects across 38 object classes spanning five typical domains (mechanics, medicine, electronics, agriculture, and food). The dataset comprises over 137.3K object instances, of which 8.6K objects are labeled as abnormal with pixel-level annotations. 
We evaluate on MANTA-Tiny, a subset of the MANTA dataset in which each class contains 800 object instances (4,000 individual images per class across 5 viewpoints), totaling 152,000 images across all 38 classes. Despite being termed ``tiny,'' it contains more images than Real-IAD. All images in MANTA-Tiny are standardized to 256×256 resolution, facilitating consistent input preprocessing since \ours processes only visual components. This contrasts with the original MANTA dataset, which requires manual cropping due to varying image dimensions and includes textual annotations that \ours does not utilize.

\begin{table*}[t!]
\centering
\resizebox{\linewidth}{!}{
\renewcommand{\arraystretch}{1}
\begin{tabular}{@{}r@{\hspace{2mm}}c@{\hspace{2mm}}c@{\hspace{2mm}}c@{\hspace{2mm}}c@{\hspace{2mm}}c@{\hspace{2mm}}c@{\hspace{2mm}}c@{\hspace{2mm}}c@{\hspace{2mm}}c@{\hspace{2mm}}c@{}}
\toprule
Method & S-AUROC & S-AP & S-F1 & I-AUROC & I-AP & I-F1 & P-AUROC & P-AP & P-F1 & P-AUPRO \\
\midrule
RD4AD $_\text{CVPR'22}$~\cite{deng2022anomaly} & 78.3 & 63.7 & 68.3 & 80.5 & 59.5 & 62.4 & 91.8 & 29.1 & 34.7 & 78.8 \\
UniAD $_\text{NeurIPS'22}$~\cite{you2022unified} & 74.9 & 59.7 & 65.2 & 82.7 & 64.3 & 65.1 & 90.7 & 26.1 & 32.4 & 79.6 \\
DeSTSeg $_\text{CVPR'23}$~\cite{zhang2023destseg} & 62.4 & 49.3 & 56.5 & 62.3 & 37.8 & 45.5 & 54.8 & 12.0 & 9.0 & 23.4 \\
RealNet $_\text{CVPR'24}$~\cite{zhang2024realnet} & 55.1 & 40.9 & 51.7 & 57.3 & 31.8 & 41.3 & 50.0 & 10.3 & 4.4 & 15.1 \\
MVAD $_\text{TMM'25}$~\cite{he2024learning} & \ul{85.0} & \ul{76.5} & \ul{73.8} & \ul{85.2} & \ul{69.9} & \ul{67.9} & \ul{93.7} & \ul{32.7} & \ul{38.5} & \ul{80.7} \\
\ourso & \textbf{88.2} & \textbf{81.9} & \textbf{78.3} & \textbf{87.0} & \textbf{73.4} & \textbf{70.5} & \textbf{94.2} & \textbf{36.8} & \textbf{41.6} & \textbf{83.1} \\
\bottomrule
\end{tabular}
}
\caption{
Average performance comparison on MANTA-Tiny dataset. The best and second-best results are marked in \textbf{bold} and \ul{underlined}.}
\label{tab:manta_tiny_average}
\end{table*}

\begin{table*}[t!]
\centering
\resizebox{\linewidth}{!}{
\renewcommand{\arraystretch}{1}
\begin{tabular}{@{}r@{\hspace{2mm}}c@{\hspace{2mm}}c@{\hspace{2mm}}c@{\hspace{2mm}}c@{\hspace{2mm}}c@{\hspace{2mm}}c@{\hspace{2mm}}c@{\hspace{2mm}}c@{\hspace{2mm}}c@{\hspace{2mm}}c@{}}
\toprule
Method & S-AUROC & S-AP & S-F1 & I-AUROC & I-AP & I-F1 & P-AUROC & P-AP & P-F1 & P-AUPRO \\
\midrule
RD4AD $_\text{CVPR'22}$~\cite{deng2022anomaly} & 85.7 & 92.4 & 87.7 & 83.0 & 79.6 & 74.4 & 97.3 & 25.8 & 33.5 & 90.4 \\
UniAD $_\text{NeurIPS'22}$~\cite{you2022unified} & 87.3 & 93.1 & 88.9 & 83.0 & 83.4 & 74.6 & 97.4 & 23.7 & 31.4 & 87.3 \\
RD++ $_\text{CVPR'23}$~\cite{tien2023revisiting} & 87.0 & 93.3 & 88.2 & 83.7 & 80.9 & 74.9 & 97.7 & 25.8 & 33.5 & 90.0 \\
SimpleNet $_\text{CVPR'23}$~\cite{liu2023simplenet} & 61.2 & 77.9 & 82.1 & 82.6 & 79.2 & 74.1 & 76.3 & 3.4 & 6.9 & 42.5 \\
DeSTSeg $_\text{CVPR'23}$~\cite{zhang2023destseg} & 89.0 & 94.6 & 89.2 & 82.3 & 79.2 & 73.2 & 94.6 & \textbf{37.9} & \textbf{41.7} & 40.6 \\
RealNet $_\text{CVPR'24}$~\cite{zhang2024realnet} & 72.0 & 87.1 & 82.5 & 65.2 & 65.4 & 62.7 & 60.5 & 27.2 & 24.9 & 28.7 \\
MambaAD $_\text{NIPS'24}$~\cite{he2024mambaad} & 85.7 & 78.3 & 74.7 & 85.7 & 72.2 & 69.1 & 93.8 & {35.5} & {40.6} & 80.7\\
ViTAD $_\text{CVIU'25}$~\cite{zhang2025exploring} & 88.8 & 94.3 & 89.3 & 82.8 & 80.0 & 73.7 & 97.1 & 23.9 & 31.8 & 84.3 \\
MVAD $_\text{TMM'25}$~\cite{he2024learning} & \ul{90.2} & \ul{95.3} & \ul{90.1} & \ul{86.6} & \ul{84.8} & \ul{77.2} & \textbf{97.9} & 30.3 & 36.8 & \ul{91.2} \\
\ourso & \textbf{91.8} & \textbf{95.9} & \textbf{91.2} & \textbf{87.5} & \textbf{85.6} & \textbf{78.2} & \ul{97.8} & \ul{35.6} & \ul{40.8} & \textbf{92.2} \\
\bottomrule
\end{tabular}
}
\caption{
Average performance comparison on Real-IAD dataset. The best and second-best results are marked in \textbf{bold} and \ul{underlined}.}
\label{tab:realiad_average}
\end{table*}


\noindent\textbf{Metrics.} Following~\cite{he2024learning}, we choose the metrics for evaluation across three levels of analysis. 
\textit{Sample-level} metrics aggregate multi-view information to determine whether an object instance is abnormal, measured by S-AUROC, S-AP (Average Precision), and S-F1 (F1 score).
\textit{Image-level} metrics evaluate the detection capability for individual images using I-AUROC, I-AP, and I-F1.
\textit{Pixel-level} metrics assess the model's ability to localize abnormal regions precisely, including P-AUROC, P-AP, P-F1, and P-AUPRO (Area Under the Per-Region-Overlap).
All results are the mean performance of all object classes for each dataset.

\begin{figure*}[t]
    \centering
    \includegraphics[width=1\textwidth]{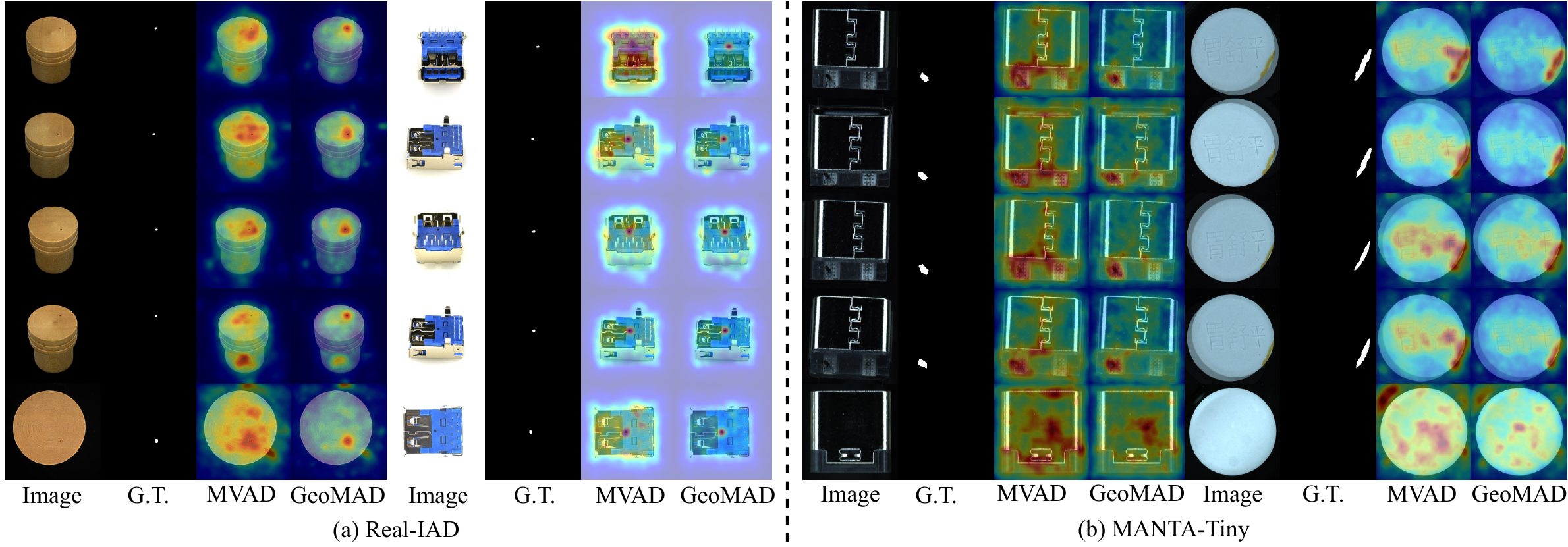}
    \vspace{-2em}
    \caption{
    Qualitative comparison on Real-IAD (left) and MANTA-Tiny (right). Each row shows one view of the same object instance ($V=5$ views per sample). \ours yields more spatially coherent localization with fewer false activations. }
    \label{fig:vis_realiad_manta}
    \vspace{-1em}
\end{figure*}

\noindent\textbf{Implementation details.} All input images from both the MANTA-Tiny and Real-IAD datasets are resized to 256$\times$256 pixels without additional data augmentation. In the multi-view setting, we utilize all available view images with 8 input samples per batch, yielding a total batch size of 40. 
Following RD4AD~\cite{deng2022anomaly}, we employ a teacher-student architecture in which the teacher network is a pre-trained ResNet-34 on ImageNet-1K that remains frozen during training, while the student network uses the same ResNet-34 architecture to learn to mimic the teacher's feature representations. 
Training is conducted for 50 epochs on a single NVIDIA RTX 5090 GPU. To ensure fair comparison, we reimplement prior methods on both the Real-IAD and MANTA-Tiny datasets in unified settings.

\subsection{Comparison with the SOTA methods}

\SC{
We compare \ours against a broad set of recent AD methods covering four design paradigms:
reconstruction-based RD4AD~\cite{deng2022anomaly} and RD++~\cite{tien2023revisiting}, 
unified frameworks UniAD~\cite{you2022unified}, MambaAD~\cite{he2024mambaad}, and ViTAD~\cite{zhang2025exploring}, 
feature-based SimpleNet~\cite{liu2023simplenet} and RealNet~\cite{zhang2024realnet}, 
segmentation-based DeSTSeg~\cite{zhang2023destseg}, 
and the multi-view method MVAD~\cite{he2024learning}.
For fair comparison, all methods are configured in the multi-class setting, training a single model across all classes.
}

\noindent\textbf{Results on Real-IAD.}
\SC{
As shown in Table~\ref{tab:realiad_average}, \ours achieves the best overall performance on Real-IAD, ranking first on 8 out of 10 metrics and second on the remaining two (P-AP and P-F1).
Compared to the strongest reconstruction-based baseline MVAD~\cite{he2024learning}, \ours improves S-AUROC by 1.6, P-AP by 5.3, and P-F1 by 4.0 points, with consistent gains across sample-, image-, and pixel-level metrics.
The only metrics on which \ours does not lead are P-AP and P-F1, where DeSTSeg~\cite{zhang2023destseg} ranks first.
We note that DeSTSeg is trained with synthetic anomalies and pseudo ground-truth masks, providing direct pixel-level supervision for its segmentation head, whereas \ours and the other baselines are trained only on anomaly-free data and derive anomaly maps from teacher--student divergence.
This supervision asymmetry favors DeSTSeg on per-pixel metrics, but the advantage does not transfer to region-level localization (P-AUPRO) or to sample- and image-level detection.
}

\noindent\textbf{Results on MANTA-Tiny.}
\SC{
The advantage of \ours grows on the more challenging MANTA-Tiny (Table~\ref{tab:manta_tiny_average}), where high appearance variability makes multi-view reasoning harder.
\ours achieves the best result across all 10 metrics, surpassing the second-best MVAD by clear margins on every level: S-AUROC +3.2, S-AP +5.4, I-AP +3.5, P-AP +4.1, and P-AUPRO +2.4.
We attribute these gains to CDFM's deformable cross-view fusion and DVA's distributional regularization, which together provide stronger inductive biases when each individual view carries limited or inconsistent evidence.
}

\noindent\textbf{Qualitative comparison.}
Figure~\ref{fig:vis_realiad_manta} shows qualitative comparisons against MVAD on both datasets.
MVAD's patch-based fusion tends to fragment defects near patch boundaries and over-activate on background regions, while \ours produces spatially coherent localization that remains consistent across the five views of the same object instance.
The behavior reflects CDFM's deformable cross-view sampling and DVA's distributional regularization, and aligns with our quantitative results: \ours achieves notably higher P-AP and P-F1 than MVAD, confirming that resolving patch-boundary fragmentation directly reduces false positive predictions.
The gap is more pronounced on MANTA-Tiny, where MVAD produces diffuse activations across the object surface while \ours retains focused localization.

\subsection{Ablations}

\noindent \textbf{Component Analysis.}
\SC{
We assess the individual contributions of CDFM and DVA in Table~\ref{tab:ablation_study} by incrementally adding each component to a baseline that uses neither.
Adding CDFM yields substantial gains on both datasets, with particularly large improvements on the more challenging MANTA-Tiny benchmark.
This confirms that deformable cross-view fusion provides a strong inductive bias for multi-view AD, allowing the model to learn view correspondence directly from data without explicit 3D supervision.
Adding DVA on top of CDFM further improves performance across the majority of metrics on both datasets, with the gains being most consistent on MANTA-Tiny.
While the improvement from DVA is smaller than that of CDFM, this is expected: CDFM establishes local cross-view correspondence, leaving DVA the complementary role of regularizing the resulting bottleneck distributions to be globally view-consistent.
Together, the two components yield the best overall performance on both datasets, validating their complementary design.
}

\begin{table}[!t]
\vspace{1em}
\setlength{\tabcolsep}{4pt}
\centering
\resizebox{\linewidth}{!}
{
\begin{tabular}{@{}ccccccccccccc@{}}
\toprule
Dataset & CDFM & DVA & S-AUROC & S-AP & S-F1 & I-AUROC & I-AP & I-F1 & P-AUROC & P-AP & P-F1 & P-AUPRO \\
\midrule
\multirow{3}{*}{MANTA-Tiny} & \ccross & \ccross & 78.3 & 63.7 & 68.3 & 80.5 & 59.5 & 62.4 & 91.8 & 29.1 & 34.7 & 78.9 \\
& \ccheck & \ccross & \ul{87.8} & \ul{81.4} & \ul{77.7} & \ul{86.6} & \ul{72.7} & \ul{70.0} & \ul{94.0} & \ul{36.0} & \ul{40.9} & \ul{82.4} \\
& \ccheck & \ccheck & \textbf{88.2} & \textbf{81.9} & \textbf{78.3} & \textbf{87.0} & \textbf{73.4} & \textbf{70.5} & \textbf{94.2} & \textbf{36.8} & \textbf{41.6} & \textbf{83.1} \\
\midrule
\multirow{3}{*}{Real-IAD} & \ccross & \ccross & 84.7 & 91.8 & 87.5 & {82.0} & 78.5 & {73.7} & \ul{97.0} & 26.1 & 33.3 & {89.5} \\
& \ccheck & \ccross & \ul{91.6} & \ul{95.8} & \ul{91.0} & \ul{87.3} & \ul{85.5} & \ul{77.9} & \textbf{97.8} & \textbf{35.7} & \textbf{41.0} & \ul{92.0} \\
& \ccheck & \ccheck & \textbf{91.8} & \textbf{95.9} & \textbf{91.2} & \textbf{87.5} & \textbf{85.6} & \textbf{78.2} & \textbf{97.8} & \ul{35.6} & \ul{40.8} & \textbf{92.2} \\

\bottomrule
\end{tabular}
}
\caption{
Ablation study on CDFM and DVA.
We evaluate the contribution of Cross-view Deformable Fusion Module (CDFM) and Distributional View Alignment (DVA) on MANTA-Tiny and Real-IAD, reporting sample-level (S), image-level (I), and pixel-level (P) metrics. 
\textbf{Bold} indicates the best result and \ul{underlined} indicates the second best.
}
\vspace{-1em}
\label{tab:ablation_study}
\end{table}

\noindent\textbf{Smoothing Kernel Size Analysis.} 
\SC{
We study how the receptive field of the smoothing kernel in CDFM affects anomaly detection performance, comparing single-kernel sizes from 3 to 9 and a multi-kernel variant $\{5, 7\}$ against a no-smoothing baseline (``-''). As shown in Table~\ref{tab:ablation_kernel_size}, smoothing is beneficial when the receptive field is large enough: kernel sizes 5, 7, and 9 all outperform the no-smoothing baseline, while the smallest kernel of size 3 actually underperforms it, suggesting that an overly local average disrupts useful structure without effectively denoising boundary artifacts. Among the effective kernels, size 7 provides the best overall balance, achieving the strongest results on the majority of image- and pixel-level metrics. Larger kernels such as 9 remain competitive on sample-level metrics but slightly degrade pixel-level localization, indicating that excessive smoothing blurs fine-grained anomaly cues.
Interestingly, combining multiple kernels $\{5, 7\}$ does not yield gains and instead performs on par with kernel 3, suggesting that a single well-chosen receptive field is sufficient for the smoothing operation.
We therefore adopt kernel size 7 as the default in all subsequent experiments.
}

\begin{table}[!t]
\vspace{.5em}
\setlength{\tabcolsep}{4pt}
\centering
\resizebox{\linewidth}{!}
{
\begin{tabular}{ccccccccccc}
\toprule
Kernel Size & S-AUROC & S-AP & S-F1 & I-AUROC & I-AP & I-F1 & P-AUROC & P-AP & P-F1 & P-AUPRO \\
\midrule
- & 91.3 & 95.7 & 90.9 & 87.0 & 85.1 & 77.6 & \textbf{97.8} & 35.5 & 40.5 & 91.6 \\
\{3\} & 91.0 & 95.5 & 90.7 & 86.8 & 85.0 & 77.4 & \textbf{97.8} & 35.2 & 40.3 & 91.4 \\
\{5\} & 91.6 & 95.8 & \ul{91.0} & \ul{87.4} & \ul{85.5} & \ul{78.0} & \textbf{97.8} & \textbf{36.1} & \textbf{41.2} & \ul{92.0}  \\
\{7\} & \ul{91.8} & \ul{95.9} & \textbf{91.2} & \textbf{87.5} & \textbf{85.6} & \textbf{78.2} & \textbf{97.8} & \ul{35.6} & \ul{40.8} & \textbf{92.2} \\
\{9\} & \textbf{92.0} & \textbf{96.0} & \textbf{91.2} & 87.3 & 85.3 & 77.9 & \textbf{97.8} & 35.1 & 40.5 & \ul{92.0} \\
\{5,7\} & 91.0 & 95.5 & 90.7 & 86.8 & 84.9 & 77.4 & \textbf{97.8} & 35.4 & 40.5 & 91.4\\
\bottomrule
\end{tabular}
}
\caption{
Impact of smoothing kernel size in CDFM.
We compare single-kernel sizes \{3, 5, 7, 9\} and a multi-kernel variant \{5, 7\} against a no-smoothing baseline (denoted ``-''). 
The best and second-best results are marked in \textbf{bold} and \ul{underlined}.
}
\label{tab:ablation_kernel_size}
\vspace{-.5em}
\end{table}

\begin{table}[!t]
\setlength{\tabcolsep}{4pt}
\centering
\resizebox{\linewidth}{!}
{
\begin{tabular}{@{}ccccccccccc@{}}
\toprule
DVA Type & S-AUROC & S-AP & S-F1 & I-AUROC & I-AP & I-F1 & P-AUROC & P-AP & P-F1 & P-AUPRO \\
\midrule
View Pair &  91.6 & 95.8 & 91.2 & 86.9 & 84.8 & 77.4 & 97.8 & 34.3 & 39.8 & 91.7\\
View Centric & \textbf{91.8} & \textbf{95.9} & \textbf{91.2} & \textbf{87.5} & \textbf{85.6} & \textbf{78.2} & \textbf{97.8} & \textbf{35.6} & \textbf{40.8} & \textbf{92.2} \\
\bottomrule
\end{tabular}
}
\caption{
Alignment target in DVA.
Comparison between pairwise alignment (View Pair) and view-centric alignment toward a shared per-instance centroid (View Centric).
}
\label{tab:ablation_CVKLD}
\vspace{-1em}
\end{table}

\noindent\textbf{Impact of Missing Views.} 
\SC{
A practical concern in multi-view deployment is robustness when certain camera views become unavailable due to occlusion or hardware failure. 
To assess this, we take the model trained with the default 5 views and evaluate it with progressively fewer views (5$\rightarrow$2) at inference time. 
As shown in Table~\ref{tab:missingview}, results are reported at the Sample (S), Image (I), and Pixel (P) levels.
Image- and pixel-level performance remain highly stable as views are removed: I-AUROC drops by only 0.8 points and P-AUROC by 0.7 points even when 60\% of views are unavailable, indicating that per-view detection and localization do not depend on the presence of any specific view.
Sample-level metrics show a larger but graceful decline, which is structurally expected since the sample score aggregates evidence across views and naturally weakens as the number of views decreases.
Overall, the method retains strong per-view performance under substantial view loss, supporting its deployment in scenarios where capturing all views is not always feasible.
}

\noindent\textbf{Alignment Target in DVA.}
\SC{
A central design choice in DVA is what each view should be aligned against. We compare two natural targets: \textit{View Pair}, which enforces consistency between every pair of view distributions via symmetric KL divergence ($\mathcal{O}(V^2)$ constraints), and \textit{View Centric}, which aligns each view against a shared per-instance centroid formed from the views themselves ($\mathcal{O}(V)$ constraints, Eq.~\ref{eq:view_centroid}).
As shown in Table~\ref{tab:ablation_CVKLD}, View Centric consistently outperforms View Pair across all metrics, with the most pronounced gains on fine-grained pixel-level metrics.
Beyond the lower constraint count, the view-centric formulation avoids a conflicting-gradient pathology of pairwise objectives, where a single badly-aligned pair can dominate the gradient and pull two views apart along an axis that other pairs are simultaneously aligning.
A shared centroid target removes this contention by giving all views one symmetric reference, providing a more coherent learning signal.
}

\noindent\textbf{Computational Efficiency.}
\SC{
Table~\ref{tab:compute_resource} compares the computational cost of \ours with representative SOTA methods, measured with batch size 40. \ours maintains a compact footprint of 19.0M parameters, on par with the most parameter-efficient baseline MVAD~\cite{he2024learning}, while using 4.2$\times$ fewer parameters and 3.0$\times$ lower FLOPs than RD4AD~\cite{deng2022anomaly}. Although UniAD~\cite{you2022unified} achieves the lowest FLOPs, \ours offers a more favorable balance across all three dimensions, particularly compared to MambaAD~\cite{he2024mambaad}, whose training memory is over 3$\times$ higher than ours.
Notably, IDIF is estimated to require over 149M parameters based on its voxel-related modules and comparable architectures~\cite{zhang2023destseg, xie2019pix2vox}, underscoring the efficiency of our non-voxel design.
}

\begin{table}[t!]
\renewcommand{\arraystretch}{0.9}
\centering
\resizebox{\linewidth}{!}
{
\begin{tabular}{lcccccccccc}
\toprule
Views & S-AUROC & S-AP & S-F1 & I-AUROC & I-AP & I-F1 & P-AUROC & P-AP & P-F1 & P-AUPRO \\
\midrule
5 (default) & \textbf{91.8} & \textbf{95.9} & \textbf{91.2} & \textbf{87.5} & \textbf{85.6} & \textbf{78.2} & \textbf{97.8} & \textbf{35.6} & \textbf{40.8} & \textbf{92.2} \\
4           & 90.9 & 95.5 & 90.4 & 87.2 & 85.3 & 78.0 & 97.7 & 35.5 & \textbf{40.8} & 91.9 \\
3           & 89.1 & 94.7 & 89.1 & 86.9 & 84.8 & 77.5 & 97.5 & 35.3 & 40.6 & 91.8 \\
2           & 85.5 & 92.9 & 87.3 & 86.7 & 83.4 & 76.5 & 97.1 & 34.2 & 39.6 & 91.5 \\
\bottomrule
\end{tabular}
}
\caption{
Robustness to missing views on the Real-IAD dataset.
We evaluate model performance when reducing the number of available views from the default 5 down to 2 at inference time, using the model trained with all 5 views.
}
\label{tab:missingview}
\end{table}

\begin{table}[t!]
\renewcommand{\arraystretch}{1}
\centering
\resizebox{.7\linewidth}{!}
{
\centering
\begin{tabular}{rcccc}
\toprule
{Method} & {Parameters} & {FLOPs} & {Train Memory} \\
\midrule
RD4AD $_\text{CVPR'22}$~\cite{deng2022anomaly} & 80.61M & 142.0G & 11,614M \\
UniAD $_\text{NeurIPS'22}$~\cite{you2022unified} & 24.52M & \textbf{16.8G} & 7,324M \\
ViTAD $_\text{CVIU'25}$~\cite{zhang2025exploring} & 38.6M & 48.3G & \textbf{4,842M} \\
MambaAD $_\text{NIPS'24}$~\cite{he2024mambaad} & 25.7M & 41.5G & 28,994M \\
MVAD $_\text{TMM'25}$~\cite{he2024learning} & \textbf{18.4M} & 45.0G & 6,744M \\
\ourso & 19.0M & 46.8G & 8,310M \\
\bottomrule
\end{tabular}
}
\caption{
Computational efficiency comparison.
Parameters, FLOPs, and training memory of \ours against representative SOTA methods.
}
\label{tab:compute_resource}
\vspace{-1em}
\end{table}

\section{Conclusion}

\SC{
We propose \ours, a unified model for multi-view, multi-class anomaly detection that addresses two complementary limitations: geometric correspondence deficiency and distributional inconsistency. For cross-view geometric correspondence, we propose \textit{Cross-view Deformable Fusion Module} (CDFM), a calibration-free 2D bottleneck design that learns content-adaptive, view-pair-specific sampling offsets with image-global reach across a multi-scale window pyramid. For distributional consistency, we introduce \textit{Distributional View Alignment} (DVA), which aligns each view's bottleneck distribution against a stop-gradient view-centric target, requiring no anomaly labels, pixel-level matching, or 3D supervision. Together, CDFM and DVA combine local deformable correspondence with global distributional consistency in a single 2D architecture. Experiments on Real-IAD and MANTA-Tiny show that \ours improves unified multi-view anomaly detection and localization over existing methods, supporting the value of jointly addressing geometric and distributional gaps in cross-view learning.
Future work could explore extending \ours to handle dynamic view configurations.
}

\noindent\textbf{Acknowledgment:}
The work of Shang-Fu Chen and Wen-Huang Cheng was supported in part by the National Science and Technology Council, Taiwan (Grants: NSTC-112-2628-E-002-033-MY4 and NSTC-114-2634-F-002-004), the Taiwan Centers of Excellence (TCE), and the Center of Data Intelligence: Technologies, Applications, and Systems (Grants: 115L900901 / 115L900902 / 115L900903), National Taiwan University, from the Featured Areas Research Center Program within the framework of the Higher Education Sprout Project by the Ministry of Education, Taiwan. 
The work of Kai-Lung Hua was supported in part by the National Science and Technology Council, Taiwan (Grants: NSTC-114-2221-E-011-045-MY3 and NSTC-114-2221-E-011-041-MY3).
Kuan-Chuan Peng was exclusively supported by MERL.

\bibliography{egbib}
\end{document}